\documentclass[runningheads]{llncs}
\usepackage{graphicx}
\usepackage{amsmath,amssymb} % define this before the line numbering.
\usepackage{color}
\usepackage[width=122mm,left=12mm,paperwidth=146mm,height=193mm,top=12mm,paperheight=217mm]{geometry}
\usepackage{cite}
\usepackage{makecell}
\usepackage{enumitem}
\usepackage{threeparttable}
\usepackage{color}
\usepackage[normalem]{ulem}
\usepackage{multirow}
\usepackage{float}
\usepackage{amsfonts}
\usepackage{bm}
\usepackage{array}
\usepackage[table]{xcolor}
\usepackage{colortbl}
\usepackage[symbol]{footmisc}
\usepackage{hyperref}

\DeclareMathOperator*{\argmax}{argmax}

\newcommand{\etal}{\textit{et al}.}
\newcommand{\ie}{\textit{i}.\textit{e}.}
\newcommand{\eg}{\textit{e}.\textit{g}.}

\renewcommand{\thefootnote}{\fnsymbol{footnote}}

\makeatletter
\newcommand{\thickhline}{%
    \noalign {\ifnum 0=`}\fi \hrule height 1pt
    \futurelet \reserved@a \@xhline
}
\makeatother

\begin{document}
\title{Learning Human-Object Interactions by \\Graph Parsing Neural Networks}
% Replace with your title

\titlerunning{Graph Parsing Neural Networks (ECCV 2018)}
% Replace with a meaningful short version of your title
%
\author{Siyuan Qi$^{*1,2}$\and
Wenguan Wang$^{*1,3}$\and
Baoxiong Jia$^{1,4}$\and\\
Jianbing Shen$^{\dagger 3,5}$\and
Song-Chun Zhu$^{1,2}$}
%
%Please write out author names in full in the paper, i.e. full given and family names.
%If any authors have names that can be parsed into FirstName LastName in multiple ways, please include the correct parsing, in a comment to the volume editors:
%\index{Lastnames, Firstnames}
%(Do not uncomment it, because you may introduce extra index items if you do that, we will use scripts for introducing index entries...)
\authorrunning{S. Qi, W. Wang, B. Jia, J. Shen, S.-C. Zhu}
% Replace with shorter version of the author list. If there are more authors than fits a line, please use A. Author et al.
%

\institute{University of California, Los Angeles\and
International Center for AI and Robot Autonomy (CARA)\and
Beijing Institute of Technology\and
Peking University\and
Inception Institute of Artificial Intelligence\\
\email{syqi@cs.ucla.edu\quad
wenguanwang.ai@gmail.com\quad
baoxiongjia@ucla.edu\\
shenjianbing@bit.edu.cn\quad
sczhu@stat.ucla.edu}
}
\maketitle              % typeset the header of the contribution
\begin{abstract}
This paper addresses the task of detecting and recognizing human-object interactions (HOI) in images and videos. We introduce the Graph Parsing Neural Network (GPNN), a framework that incorporates structural knowledge while being differentiable end-to-end. For a given scene, GPNN infers a parse graph that includes i) the HOI graph structure represented by an adjacency matrix, and ii) the node labels. Within a message passing inference framework, GPNN iteratively computes the adjacency matrices and node labels. We extensively evaluate our model on three HOI detection benchmarks on images and videos: HICO-DET, V-COCO, and CAD-120 datasets. Our approach significantly outperforms state-of-art methods, verifying that GPNN is scalable to large datasets and applies to spatial-temporal settings. The code is available at \url{https://github.com/SiyuanQi/gpnn}.
\end{abstract}

\let\thefootnote\relax\footnotetext{$^{*}$ Equal contribution. $^{\dagger}$ Corresponding author.}

%%%%%%%%% BODY TEXT
\section{Introduction}
\label{sec:introduction}
The task of human-object interaction (HOI) understanding aims to infer the relationships between human and objects, such as ``riding a bike'' or ``washing a bike''. Beyond traditional visual recognition of individual instances, \textit{e.g.}, human pose estimation, action recognition, and object detection, recognizing HOIs requires a deeper semantic understanding of image contents. Recently, deep neural networks (DNNs) have shown impressive progress on above individual tasks of instance recognition, while relatively few methods  \cite{chao2015hico,chao2017learning,shenscaling,gkioxari2017detecting} were proposed for HOI recognition. This is mainly because it requires \textit{reasoning} beyond \textit{perception}, by integrating information from human, objects, and their complex relationships.

\begin{figure}[t]
\centering
\includegraphics[width=\linewidth]{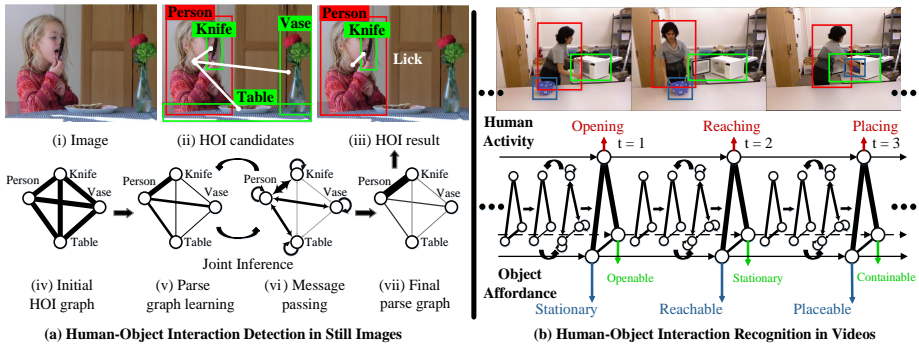}
\caption{\textbf{Illustration of the proposed GPNN for learning HOI.} GPNN offers a generic HOI representation that applies to (a) HOI detection in images and (b) HOI recognition in videos. With the integration of graphical model and neural network, GPNN can iteratively learn/infer the graph structures (a.v) and message passing (a.vi). The final parse graph explains a given scene with the graph structure (\eg, the link between the person and the knife) and the node labels (\eg, lick). A thicker edge corresponds to stronger information flow between nodes in the graph.}
\label{fig:overview}
%\vspace{-10pt}
\end{figure}

In this paper, we propose a novel model, Graph Parsing Neural Network (GPNN), for HOI recognition. The proposed GPNN offers a general framework that explicitly represents HOI structures with graphs and automatically parses the optimal graph structures in an end-to-end manner. In principle, it is an generalization of Message Passing Neural Network (MPNN)~\cite{gilmer2017neural}. An overview of GPNN is shown in \autoref{fig:overview}. The following two aspects motivate our design.

First, we seek a unified framework that utilizes the learning capability of neural networks and the power of graphical representations. Recent deep learning based HOI models showed promising results, but few touched how to interpret well and explicitly leverage spatial and temporal dependencies and human-object relations in such structured task. Aiming for this, we introduce GPNN. It inherits the complementary strengths of neural networks and graphical models, for forming a coherent HOI representation with strong learning ability. Specifically, with the structured representation of an HOI graph, the rich relations are explicitly utilized, and the information from individual elements can be efficiently integrated and broadcasted over the structures. The whole model and message passing operations are well-defined and fully differentiable. Thus it can be efficiently learned from data in an end-to-end manner.

Second, based on our efficient HOI representation and learning power, GPNN applies to diverse HOI tasks in both static and dynamic scenes. Previous studies for HOI achieved good performance in their specific domains (spatial~\cite{chao2017learning,gkioxari2017detecting} or temporal~\cite{jain2016structural,qi2017predicting,qi2018generalized}). However, none of them addresses a generic framework for representing and learning HOI in both images and videos. The key difficulty lies in the diverse relations between components. Given a set of human and objects candidates, there may exist an uncertain number of human-object interaction pairs (see \autoref{fig:overview} (a.ii) as an example). The relations become more complex after taking temporal factors into consideration. Thus pre-fixed graph structures, as adopted by most previous graphical or structured DNN models~\cite{koppula2016anticipating,jain2016structural,Wang_2018_CVPR,Fang2018}, are not an optimal choice. Seeking a better generalization ability, GPNN incorporates an essential \textit{link function} for addressing the problem of graph structure learning. It learns to infer the adjacency matrix in an end-to-end manner and thus can infer a parse graph that explicitly explains the HOI relations. With such learnable graph structure, GPNN could also limit the information flow from irrelevant nodes while encouraging message to propagate between related nodes, thus improving graph parsing.

We extensively evaluate the proposed GPNN on three HOI datasets, namely HICO-DET~\cite{chao2017learning}, V-COCO~\cite{gupta2015visual} and CAD-120~\cite{koppula2016anticipating}, for HOI detection from images (HICO-DET, V-COCO) and HOI recognition and anticipation in spatial-temporal settings (CAD-120). The experimental results verify the generality and scalability of our GPNN based HOI representation and show substantial improvements over state-of-the-art approaches, including pure graphical models and pure neural networks. We also demonstrate GPNN outperforms its variants and other graph neural networks with pre-fixed structures.

This paper makes three major contributions. \textbf{First}, we propose the GPNN that incorporates structural knowledge and DNNs for learning and inference. \textbf{Second}, with a set of well defined modular functions, GPNN addresses the HOI problem by jointly performing graph structure inference and message passing. \textbf{Third}, we empirically show that GPNN offers a scalable and generic HOI representation that applies to both static and dynamic settings.

\section{Related Work}
\label{sec:related_work}
\noindent\textbf{Human-Object Interaction.} Reasoning human actions with objects (like ``playing baseball'', ``playing guitar''), rather than recognizing individual actions (``playing'') or object instances (``baseball'', ``guitar''), is essential for a more comprehensive understanding of what is happening in the scene. Early work in HOI understanding studied Bayesian model~\cite{gupta2007objects,gupta2009observing}, utilized contextual relationship between human and objects~\cite{yao2010grouplet,yao2010modeling,yao2011human}, learned structured representations with spatial interaction and context~\cite{delaitre2011learning}, exploited compositional models~\cite{desai2012detecting}, or referred to a set of HOI exemplars~\cite{hu2013recognising}. They were mainly based on handcrafted features (\eg, color, HOG, and SIFT) with object and human detectors. More recently, inspired by the notable success of deep learning and the availability of large-scale HOI datasets~\cite{chao2015hico,chao2017learning}, several deep learning based HOI models were then proposed. Specifically, Mallya \etal~\cite{mallya2016learning} modified Fast RCNN model \cite{girshick2015fast} for HOI recognition, with the assistance of Visual Question Answering (VQA). In~\cite{shenscaling}, zero-shot learning was applied for addressing the long-tail problem in HOI recognition. In~\cite{chao2017learning}, the human proposals, object regions, and their combinations were fed into a multi-stream network for tackling the HOI detection problem. Gkioxari \etal~\cite{gkioxari2017detecting} estimated an action-type specific density map for identifying the interacted object locations, with a modified Faster RCNN architecture~\cite{ren2015faster}.

Although promising results were achieved by above deep HOI models, we still observe two unsolved issues. First, they lack a powerful tool to represent the structures in HOI tasks explicitly and encodes them into modern network architectures efficiently. Second, despite the successes in specific tasks, a complete and generic HOI representation is missing. These approaches can not be easily extended to HOI recognition from videos. Aiming to address those issues, we introduce GPNN for imposing high-level relations into DNN, leading to a powerful HOI representation that is applicable in both static and dynamic settings.

\noindent\textbf{Neural Networks with Graphs/Graphical Models.}
%Deep learning methods are able to learn flexible data representations, but lack intuitive high-level structures. Alternatively, graphical models are powerful to build structured representations, but they often need significant feature engineering.
In the literature, some approaches were proposed to combine graphical models and neural networks. The most intuitive approach is to build graphical models upon DNN, where the network that generates features is trained first, and its output is used to compute potential functions for the graphical predictor. Typical methods were used in human pose estimation~\cite{tompson2014joint}, human part parsing~\cite{xia2016pose,park2017attribute}, and semantic image segmentation~\cite{chen2016deeplab,chen2015learning}. These methods lack a deep integration in the sense that the computation process of graphical models cannot be learned end-to-end.
Some attempts~\cite{zheng2015conditional,wu2016deep,monti2016geometric,kipf2017semi,simonovsky2017dynamic,defferrard2016convolutional,niepert2016learning,seo2016structured} were made to generalize neural network operations (\eg, convolutions) directly from regular grids (\eg, images) to graphs. For the HOI problem, however, a structured representation is needed to capture the high-level spatial-temporal relations between humans and objects. Some other work integrated network architectures with graphical models~\cite{jain2016structural,gilmer2017neural} and gained promising results on applications such as scene understanding~\cite{xu2017scene, marino2016more, li2017situation}, object detection and parsing~\cite{liang2016semantic, yuan2017temporal}, and VQA~\cite{teney2016graph}. However, these methods only apply to problems that have pre-fixed graph structures. Liang \etal~\cite{liang2017interpretable} merged graph nodes using Long Short-Term Memory (LSTM) for human parsing problem, under the assumption that the nodes are mergeable.

Those methods achieved promising results in their specific tasks and well demonstrated the benefit in completing deep architectures with domain-specific structures. However, most of them are based on pre-fixed graph structures, and they have not yet been studied in HOI recognition.
In this work, we extend previous graphical neural networks with learnable graph structures, which well addresses the rich and high-level relations in HOI problems. The proposed GPNN can automatically infer the graph structure and utilize that structure for enhancing information propagation and further inference. It offers a generic HOI representation for both spatial and spatial-temporal settings. To the best of our knowledge, this is a first attempt to integrate graph models with neural networks in a unified framework to achieve state-of-art results in HOI recognition.

\begin{figure*}[t]
\centering
\includegraphics[width=\linewidth]{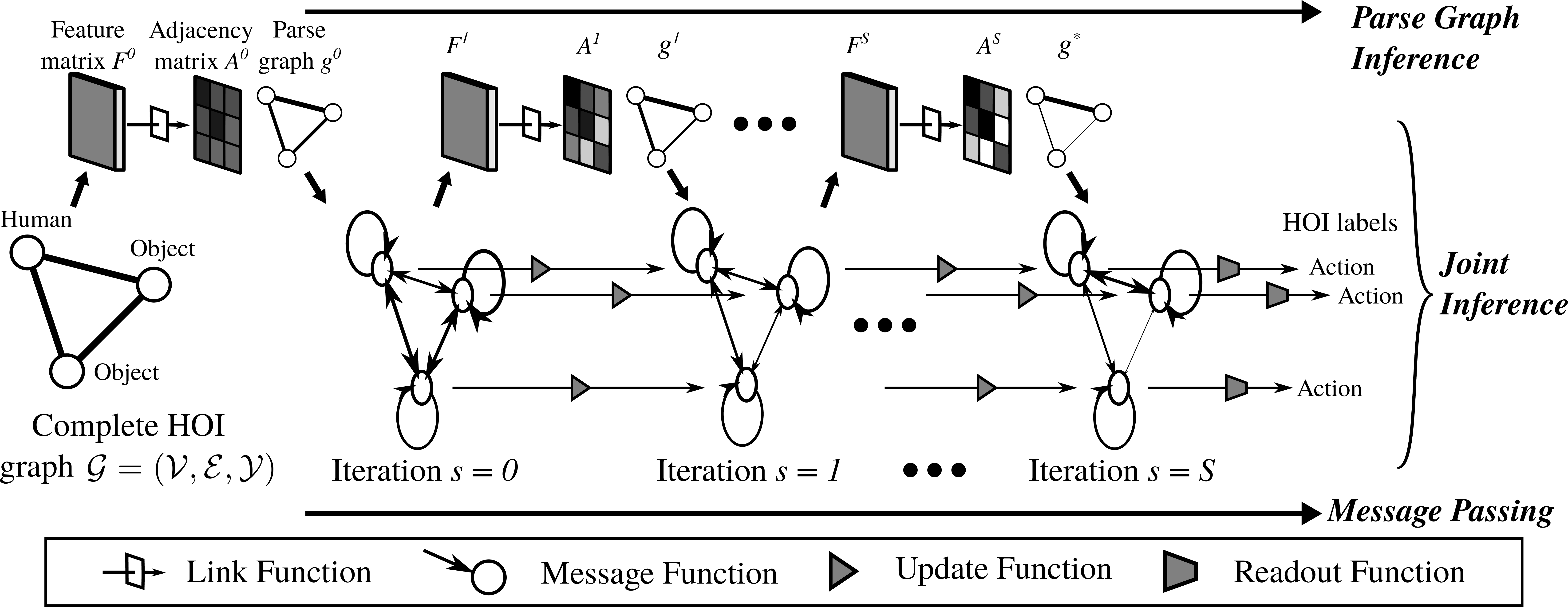}
\caption{\textbf{Illustration of the forward pass of GPNN.} GPNN takes node and edge features as input, and outputs a parse graph in a message passing fasion. The structure of the parse graph is given by a soft adjacency matrix. It is computed by the \textit{link function} based on the features (or hidden node states). The darker the color in the adjacency matrix, the stronger the connectivity is. Then \textit{message functions} compute incoming messages for each node as a weighted sum of the messages from other nodes. Thicker edges indicate larger information flows. The \textit{update functions} update the hidden internal states of each node. Above process is repeated for several steps, iteratively and jointly learning the computation of graph structures and message passing. Finally, for each node, the \textit{readout functions} output HOI action or object labels from the hidden node states. See \autoref{sec:gpnn} for more details.}
\label{fig:gpnn}
%\vspace{-10pt}
\end{figure*}

\section{Graph Parsing Neural Network for HOI}
\label{sec:gpnn}

\subsection{Formulation}
\label{sec:gpnn_forward_pass}
For HOI understanding, human and objects are represented by nodes, and their relations are defined as edges. Given a complete HOI graph that includes all the possible relationships among human and objects, we want to automatically infer a parse graph by keeping the meaningful edges and labeling the nodes.

Formally, let $\mathcal{G} = (\mathcal{V}, \mathcal{E}, \mathcal{Y})$ denote the complete HOI graph. Nodes $v \in \mathcal{V}$ take unique values from $\{1, \cdots, |\mathcal{V}|\}$. Edges $e \in \mathcal{E}$ are two-tuples $e = (v, w) \in \mathcal{V} \times \mathcal{V}$. Each node $v$ has a output state $y_{v} \in \mathcal{Y}$ that takes a value from a set of labels $\{1, \cdots, Y_{v}\}$ (\eg, actions). A parse graph $g = (\mathcal{V}_{g}, \mathcal{E}_{g}, \mathcal{Y}_{g})$ is a sub-graph of $\mathcal{G}$, where $\mathcal{V}_{g} \subseteq \mathcal{V}$ and $\mathcal{E}_{g} \subseteq \mathcal{E}$. Given the node features $\Gamma^{\mathcal{V}}$ and edge features $\Gamma^{\mathcal{E}}$, we want to infer the optimal parse graph $g^{*}$ that best explains the data according to a probability distribution $p$:
\begin{equation}
\begin{aligned}
g^{*} & = \argmax_{g}~p(g | \Gamma, \mathcal{G})
 = \argmax_{g}~p(\mathcal{V}_{g}, \mathcal{E}_{g}, \mathcal{Y}_{g} | \Gamma, \mathcal{G}) \\
& = \argmax_{g}~p(\mathcal{Y}_{g} | \mathcal{V}_{g}, \mathcal{E}_{g}, \Gamma) p(\mathcal{V}_{g}, \mathcal{E}_{g} | \Gamma, \mathcal{G})
\end{aligned}
\label{eqn:parsing_posterior}
\end{equation}
where $\Gamma = \{\Gamma^{\mathcal{V}}, \Gamma^{\mathcal{E}}\}$. Here $p(\mathcal{V}_{g}, \mathcal{E}_{g} | \Gamma, \mathcal{G})$ evaluates the graph structure, and $p(\mathcal{Y}_{g} | \mathcal{V}_{g}, \mathcal{E}_{g}, \Gamma)$ is the labeling probability for the nodes in the parse graph.
%Here we determine the nodes $\mathcal{V}_{g}$ in the parse graph $g$ by the edges $\mathcal{E}_{g}$ (\ie, nodes that have neighbors are kept) that are independent from their labels or the features.

This formulation provides us a principled guideline for designing the GPNN. We design the network to approximate the computations of $\argmax_{g} p(\mathcal{V}_{g}, \mathcal{E}_{g} | \Gamma, \mathcal{G})$ and $\argmax_{g} p(\mathcal{Y}_{g} | \mathcal{V}_{g}, \mathcal{E}_{g}, \Gamma)$. We introduce four types of functions as individual modules in the forward pass of a GPNN: \textit{link functions}, \textit{message functions}, \textit{update functions}, and \textit{readout functions} (as illustrated in \autoref{fig:gpnn}).
The link functions $L(\cdot)$ estimate the graph structure, giving an approximation of $p(\mathcal{V}_{g}, \mathcal{E}_{g} | \Gamma, \mathcal{G})$. The message, update and readout functions together resemble the belief propagation process and approximate $\argmax_{\mathcal{Y}_{g}} p(\mathcal{Y}_{g} | \mathcal{V}_{g}, \mathcal{E}_{g}, \Gamma)$.

Specifically, the link function  (\includegraphics[scale=0.08]{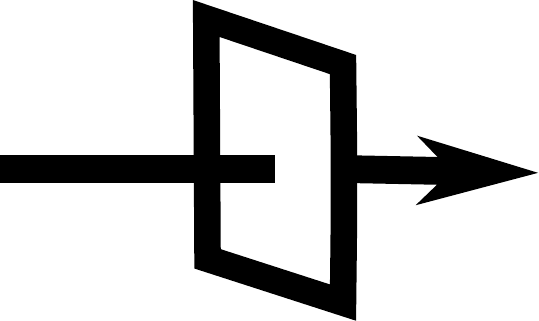}) takes edge features (\includegraphics[scale=0.04]{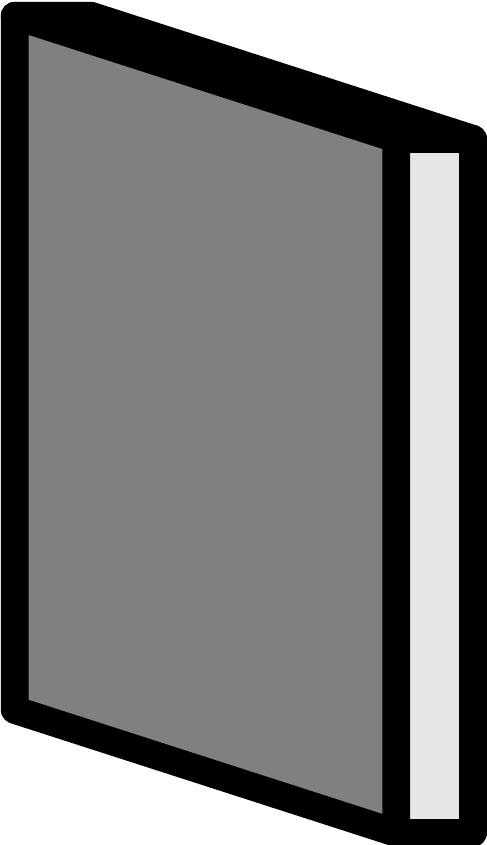}) as input and infers the connectivities between nodes. The soft adjacency matrix (\includegraphics[scale=0.04]{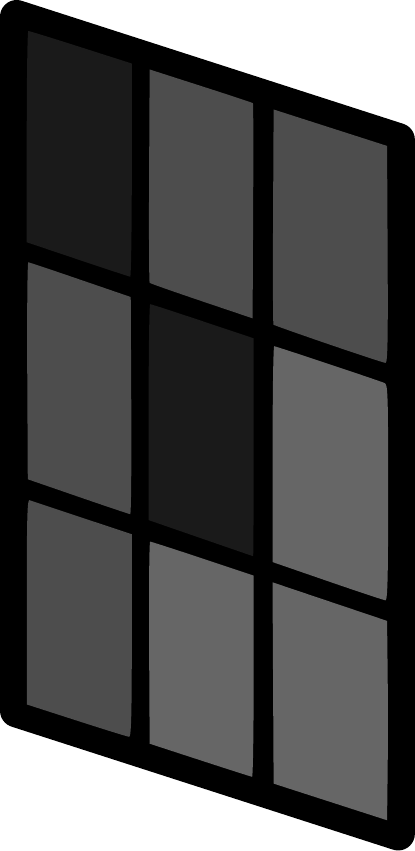}) is thus constructed and used as weights for messages passing through edges between nodes. The incoming messages for a node are summarized by the message function (\includegraphics[scale=0.06]{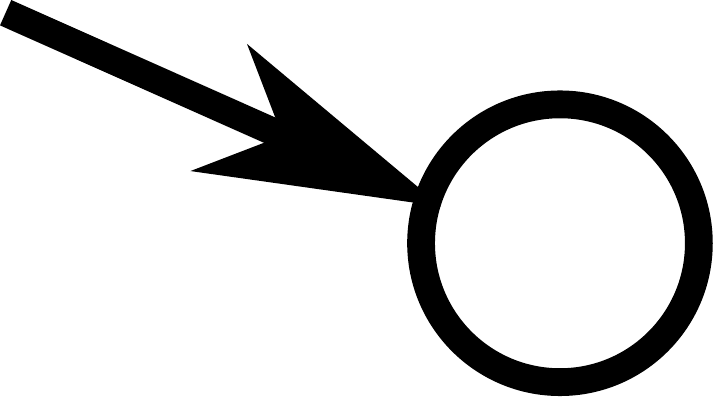}), then the hidden embedding state of the node is updated based on the messages by an update function (\includegraphics[scale=0.08]{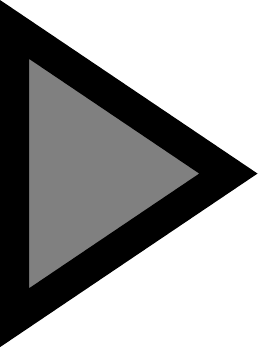}). Finally, readout functions (\includegraphics[scale=0.08]{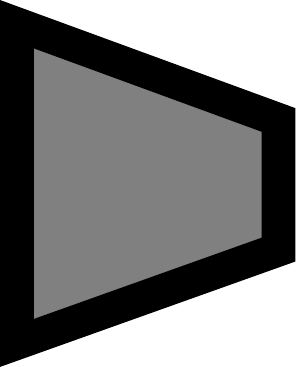}) compute the target outputs for each nodes. Those four types of functions are defined as follows:

%In principle, the forward pass of a GPNN is composed of f
%and \textit{update functions} that resemble the belief propagation algorithm~\cite{pearl1982reverend}, and \textit{readout functions} that compute the final output. The forward pass of a GPNN is illustrated in Fig.~\ref{fig:gpnn} and described as follows.

\noindent\textbf{Link Function.} %Starting with potential nodes and edges, we infer the soft connectivities between nodes from the edge features, \ie, an adjacency matrix. The adjacency matrix thus gives a graph structure, corresponding to the parse graph $g$. Specially, the link functions $L(\cdot)$ take the node features $\Gamma^{\mathcal{V}}$, and edge features $\Gamma^{\mathcal{E}}$ as input and output an adjacency matrix $A\in[0,1]^{|\mathcal{V}| \times |\mathcal{V}|}$:
We first infer an adjacency matrix that represents connectivities (\ie, the graph structure) between nodes by a link function. A link function $L(\cdot)$ takes the node features $\Gamma^{\mathcal{V}}$, and edge features $\Gamma^{\mathcal{E}}$ as input and outputs an adjacency matrix $A\in[0,1]^{|\mathcal{V}| \times |\mathcal{V}|}$:

\begin{equation}
\begin{aligned}
A_{vw} = L(\Gamma_{v}, \Gamma_{w}, \Gamma_{vw})
\end{aligned}
\label{eqn:link_function}
\end{equation}
where $A_{vw}$ denotes the $(v, w)$-th entry of the matrix $A$. Here we overload the notation and let $\Gamma_{v}$ denote node features and $\Gamma_{vw}$ denote edge features.
In this way, the structure of a parse graph $g$ can be approximated by the adjacency matrix. Then we start to propagate messages over the parse graph, where the soft adjacency matrix controls the information to be passed through edges.

\noindent\textbf{Message and Update Functions.} Based on the learned graph structure, the message passing algorithm is adopted for inference of node labels. During belief propagation, the hidden states of the nodes are iteratively updated by communicating with other nodes. Specially, message functions $M(\cdot)$ summarize messages to nodes coming from other nodes, and update functions $U(\cdot)$ update the hidden node states according to the incoming messages. At each iteration step $s$, the two functions computes:%= \sum\nolimits_{w} m_{vw}^{s}
\begin{equation}
\begin{aligned}
m_{v}^{s} = \sum\nolimits_{w} A_{vw} M(h_{v}^{s-1}, h_{w}^{s-1}, \Gamma_{vw})
\end{aligned}
\label{eqn:message_function}
\end{equation}
\begin{equation}
\begin{aligned}
h_{v}^{s} = U(h_{v}^{s-1}, m_{v}^{s})
\end{aligned}
\label{eqn:update_function}
\end{equation}
where $m_{v}^{s}$ is the summarized incoming message for node $v$ at $s$-th iteration and $h_{v}^{s}$ is the hidden state for node $v$. The node connectivity $A$ encourages the information flow between nodes in the parse graph. The message passing phase runs for $S$ steps towards convergence. At the first step, the node hidden states $h_{v}^{0}$ are initialized by node features $\Gamma_{v}$.

\noindent\textbf{Readout Function.} Finally, for each node, hidden state is fed into a readout function to output a label:
\begin{equation}
\begin{aligned}
y_{v} = R(h_{v}^{S}).
\end{aligned}
\label{eqn:readout_function}
\end{equation}
Here the readout function $R(\cdot)$ computes output $y_v$ for node $v$ by activating its hidden state $h_{v}^{S}$ (node embeddings).

\noindent\textbf{Iterative Parsing.} Based on the above four functions, the messages are passed along the graph and weighted by the learned adjacency matrix $A$. We further extend above process into a joint learning framework that iteratively infers the graph structure and propagates the information to infer node labels. In particular, instead of learning $A$ only at the beginning, we iteratively infer $A$ with the updated node information and edge features at each step $s$:
\begin{equation}
\begin{aligned}
A_{vw}^{s} = L(h_{v}^{s-1}, h_{w}^{s-1}, m_{vw}^{s-1}).
\end{aligned}
\label{eqn:link_function_iterative}
\end{equation}
Then the messages in \autoref{eqn:message_function} are redefined as:
%Then the message functions $M(\cdot)$ in Eq. \ref{eqn:message_function} is rewritten as:= \sum\nolimits_{w} m_{vw}^{s}
\begin{equation}
\begin{aligned}
m_{v}^{s}  =  \sum\nolimits_{w} A_{vw}^{s} M(h_{v}^{s-1}, h_{w}^{s-1}, \Gamma_{vw}).
\end{aligned}
\label{eqn:message_function_iterative}
\end{equation}
In this way, both the graph structure and the message update can be jointly and iteratively learned in a unified framework. In practice, we find such a strategy would bring better performance (detailed in \autoref{sec:abastudy}).

%In principle, the above process is an approximation of Eq.~\ref{eqn:parsing_posterior}.
In next section, we show that by implementing each function by neural networks, the entire system is differentiable end-to-end. Hence all the parameters can be learned using gradient-based optimization.

\subsection{Network Architecture}
\label{sec:net_ar}
%In this section, we discuss detailed implementations of above functions. %For the rest of the paper, we use $n$ to denote the number of nodes in the complete graph $G$, and $d_{v}$ and $d_{e}$ to denote the dimension of the node features and edge features respectively. %Our described framework focuses on undirected graphs, but we note that it is easy to be extended to directed graphs. It is also possible to define a readout function for a graph-level output.

\noindent\textbf{Link Function.} Given the complete HOI graph $\mathcal{G} = (\mathcal{V}, \mathcal{E}, \mathcal{Y})$, we use $d_{V}$ and $d_{E}$ to denote the dimension of the node features and the edge features, respectively. In a message passing step $s$, we first concatenate all the node features (hidden states) $\{h^s_v\in \mathbb{R}^{d_{V}}\}_v$ and all the edge features (messages) $\{m^s_{vw}\in \mathbb{R}^{d_{E}}\}_{v,w}$ to form a feature matrix $F^{s}\in \mathbb{R}^{|V| \times |V| \times (2d_{V}+d_{E})}$ (see \includegraphics[scale=0.04]{figs/feature.pdf} in \autoref{fig:gpnn}). The link function is defined as a small neural network with one or several convolutional layer(s) (with $1\times1\times(2d_{V}+d_{E}$) kernels) and a $sigmoid$ activation. Then the adjacency matrix $A^s \in[0,1]^{|\mathcal{V}| \times |\mathcal{V}|}$ can be computed as:
\begin{equation}
\begin{aligned}
A^{s} = \sigma(\mathbf{W}^L\ast F^{s}),
\end{aligned}
\label{eqn:link_imple}
\end{equation}
where $\mathbf{W}^L$ is the learnable parameters of the link function network $L(\cdot)$ and $\ast$ denotes conv operation. The $sigmoid$ operation $\sigma(\cdot)$ is for normalizing the values of the elements of $A^{s}$ into $[0,1]$. The essential effect of multiple convolutional layers with $1 \times 1$ kernels is similar to fully connected layers applied to each individual edge features, except that the filter weights are shared by all the edges. In practice, we find such operation generates good enough results and leads to a high computation efficiency.

%\noindent\textbf{Link Function.}
%An intuitive design of link functions would be linear transformations, \ie, $L(h_{v}^{s}, h_{w}^{s}, m_{vw}^{s-1}) = C \cdot [h_{v}^{s}, h_{w}^{s}, m_{vw}^{s-1}] + b$, where $C$ and $b$ are learnable parameters, and $[.,.]$ denotes concatenation. However, scalability would be an issue for this implementation. The time complexity for computing an adjacency matrix is $O(n^2)$ in this case.
%
%To address this issue we propose to organize the features into a $n \times n \times (2d_{v}+d_{e})$ feature matrix $\Gamma_{A}$ and directly apply $1 \times 1$ convolution filters to it. The essential effect of multiple convolutional layers with $1 \times 1$ kernels is similar to fully connected layers applied to each individual edge features, except that the filter weights are shared by all the edges. Hence the link function is $L(h_{v}^{s}, h_{w}^{s}, m_{vw}^{s-1}) = f(\Gamma_{A})$, where $f(\cdot)$ is a neural network with multiple convolution layers, and $\Gamma_{A}$ is the features matrix with $(\Gamma_{A})_{v, w} = [\Gamma_{v}, \Gamma_{w}, \Gamma_{vw}]$. This reduces the time complexity by $n^2$ times.

For spatial-temporal problems where the adjacency matrices should account for the previous states, we use convolutional LSTMs~\cite{xingjian2015convolutional} for modeling $L(\cdot)$ in temporal domain. At time $t$, the link function takes $F^{s,t}$ as input features and the previous adjacency matrix $A^{s, t-1}$ as hidden state: $A^{s,t} = convLSTM(F^{s,t}, A^{s, t-1})$.
%\begin{equation}
%\begin{aligned}
%A^{s,t} = convLSTM(\underbrace{A^{S, t-1}}_{hidden~state}, ~~\underbrace{F^{s,t}}_{feature}).
%\end{aligned}
%\end{equation}
Again, the kernel size for the conv layer in convLSTM is $1 \times 1 \times (2d_{V}+d_{E})$.

\noindent\textbf{Message Function.}
In our implementation, the message function $M(\cdot)$ in \autoref{eqn:message_function} is computed by:
\begin{equation}
\begin{aligned}
M(h_{v}, h_{w}, \Gamma_{vw}) = [\mathbf{W}_V^M h_{v}, \mathbf{W}_V^M h_{w}, \mathbf{W}_E^M \Gamma_{vw}],
\end{aligned}
\end{equation}
where $[.,.]$ denotes concatenation. It concatenates the outputs of linear transforms (\ie, fully connected layers parametrized by $\mathbf{W}_V^M$ and $\mathbf{W}_E^M$) that takes node hidden states $h_{v}$ or edge features $\Gamma_{vw}$ as input. %Camera-ready: The message functions can be easily extended to handle problems that require different node types and different edge types. One simple but efficient solution would be putting different types of features into different positions of the entire feature vector. %We note that the message functions can be easily extended to handle problems that require different node types and different edge types. One simple but efficient solution would be putting different types of features into different positions of the entire feature vector.

\noindent\textbf{Update Function.} Recurrent neural networks~\cite{elman1990finding,hochreiter1997long} are natural choices for simulating the iterative update process, as done by previous works \cite{gilmer2017neural}. Here we apply Gated Recurrent Unit (GRU)~\cite{cho2014properties} as the update function, because of its recurrent nature and smaller amount of parameters. Thus the update function in \autoref{eqn:update_function} is implemented as:
\begin{equation}
\begin{aligned}
h_{v}^{s} = U(h_{v}^{s-1}, m_{v}^{s}) = GRU(h_{v}^{s-1}, m_{v}^{s}),
\end{aligned}
\end{equation}
where $h_{v}^{s}$ is the hidden state and $m_{v}^{s}$ is used as input features. As demonstrated in \cite{li2015gated}, the GRU is more effective than vanilla recurrent neural networks.
%Linear function is also an option for the update functions, as illustrated by Scarselli \etal~\cite{scarselli2009graph}. However, to ensure convergence the parameters must be constrained so that each update step is a contraction map. Li \etal~\cite{li2015gated} proposed Gated Graph Neural Networks, in which the update function is essentially a Gated Recurrent Unit (GRU)~\cite{cho2014properties}. In practice, it is more effective than vanilla recurrent neural networks.

\noindent\textbf{Readout Function.}
A typical implementation of readout functions is combining several fully connected layers (parameterized by $\mathbf{W}^R$) followed by an activation function:
\begin{equation}
\begin{aligned}
y_v = R(h_v^S) = \varphi(\mathbf{W}^R h_v^S).
\end{aligned}
\end{equation}
Here the activation function $\varphi(\cdot)$ can be used as \textit{softmax} (one-class outputs) or \textit{sigmoid} (multi-class outputs) according to different HOI tasks.

In this way, the entire GPNN is implemented to be fully differentiable and end-to-end trainable. The loss for specific HOI task can be computed for the outputs of readout functions, and the error can propagate back according to chain rule. In next section, we will offer more details for implementing GPNN for HOI tasks on spatial and spatial-temporal settings and present qualitative as well as quantitative results.

\section{Experiments}
%The proposed GPNN model offers a unified and HOI representation.
To verify the effectiveness and generic applicability of GPNN, we perform experiments on two HOI problems: i) HOI detection in images \cite{chao2017learning,gupta2015visual}, and ii) HOI recognition and anticipation from videos~\cite{koppula2016anticipating}. The first experiment is performed on HICO-DET~\cite{chao2017learning} and V-COCO~\cite{gupta2015visual} datasets, showing that our approach is scalable to large datasets (about 60K images in total) and achieves a good detection accuracy over a large number of classes (more than 600 classes of HOIs). The second experiment is reported on CAD-120 dataset~\cite{koppula2016anticipating}, showing that our method is well applicable to spatial-temporal domains.

\subsection{Human-Object Interaction Detection in Images}
\label{sec:humanobjectinter}
For HOI detection in an image, the goal is to detect pairs of a human and an object bounding box with an interaction class label connecting them. %HOI recognition goes beyond traditional object recognition or detection tasks. It requires a deeper understanding of the context of humans and objects.

%%%%%%%%%%%%%%%%%%%%%%%%%%%%%Table 2%%%%%%%%%%%%%%%%%%%%%%%%%%%%%%%%%%%%%%%%%%%%
\begin{table}[!ht]%[!htbp]
\caption{\textbf{HOI detection results (mAP) on HICO-DET dataset~\cite{chao2017learning}}. Higher values are better. The best scores are marked in \textbf{bold}.
%See \autoref{sec:humanobjectinter} for more details
}
\label{tab:hico_result}
\resizebox{\textwidth}{!}{
\setlength\tabcolsep{5pt}
\renewcommand\arraystretch{1.1}
\begin{tabular}{l||c||c||c}  % {lccc}
\hline\thickhline
Methods & Full (mAP \%) $\uparrow$ & Rare (mAP \%) $\uparrow$ & Non-rare (mAP \%) $\uparrow$ \\
\hline
\hline
 Random & $1.35\times10^{-3}$ & $5.72\times10^{-4}$ & $1.62\times10^{-3}$ \\
 Fast-RCNN(union)~\cite{girshick2015fast} & 1.75 & 0.58 & 2.10 \\
 Fast-RCNN(score)~\cite{girshick2015fast} & 2.85 & 1.55 & 3.23 \\
 HO-RCNN~\cite{chao2017learning} & 5.73 & 3.21 & 6.48 \\
 HO-RCNN+IP~\cite{chao2017learning} &7.30 &4.68 &8.08\\
 HO-RCNN+IP+S~\cite{chao2017learning} & 7.81 & 5.37 & 8.54 \\
 Gupta \etal \cite{gupta2015visual} &9.09 &7.02 &9.71\\
 Shen \etal \cite{shenscaling} &6.46 &4.24 &7.12\\
 InteractNet \cite{gkioxari2017detecting} &9.94 &7.16 &10.77\\
 \hline
\textbf{GPNN} & \textbf{13.11} & \textbf{9.34} & \textbf{14.23} \\
%\textit{Performance Gain} 	& +6.15 & +3.39 & +5.38 \\
\textit{Performance Gain}(\%) 	& 31.89 & 30.45 & 32.13 \\
\hline
\end{tabular}
}
\end{table}

\begin{figure*}[t]
\centering
\includegraphics[width=\linewidth]{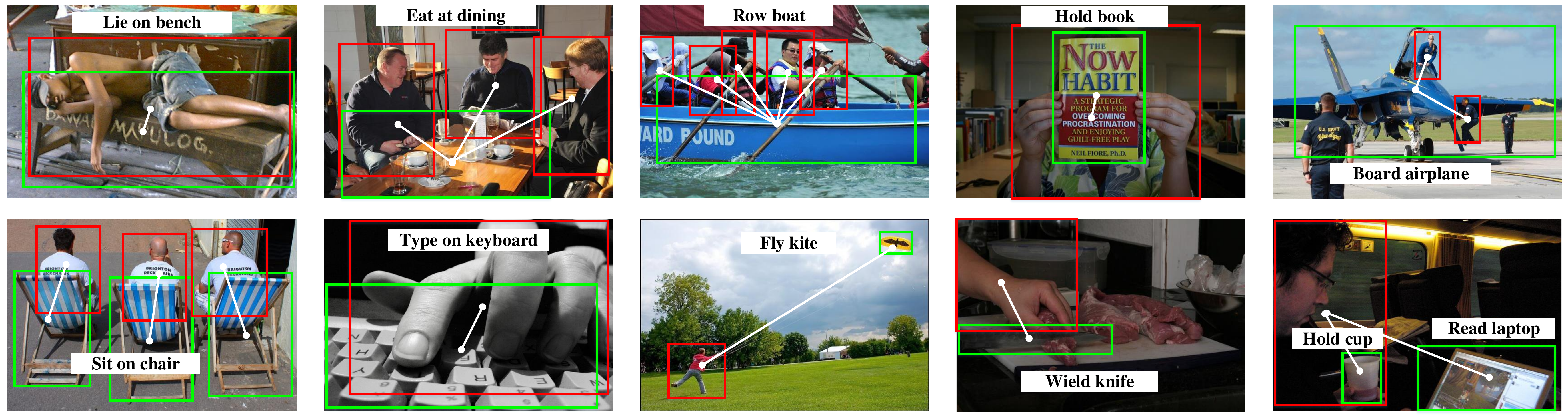}
\caption{\textbf{HOI detection results on HICO-DET~\cite{chao2017learning} test images.} Human and objects are shown in red and green rectangles, respectively. Best viewed in color.}
\label{fig:hico_results}
%\vspace{-10pt}
\end{figure*}

\noindent\textbf{Datasets.} We use HICO-DET~\cite{chao2017learning} and V-COCO~\cite{gupta2015visual} datasets for benchmarking our GPNN model. HICO-DET provides more than 150K annotated instances of human-object pairs in 47,051 images (37,536 training and 9,515 testing). %It covers the 600 HOI categories in HICO~\cite{chao2015hico}, which is an image classification dataset based on HOI categories.
It has the same 80 object categories as MS-COCO~\cite{lin2014microsoft} and 117 action categories. V-COCO is a subset of MS-COCO~\cite{lin2014microsoft}. It consists of a total of 10,346 images with
16,199 people instances, where $\sim$2.5K images in the train set, $\sim$2.8K images for validation and $\sim$4.9K images for testing. Each annotated person has binary
labels for 26 different action classes. Note that three actions (\ie, \emph{cut}, \emph{eat}, and \emph{hit}) are annotated with two types of targets: \textit{instrument} and \textit{direct object}.

\noindent\textbf{Implementation Details.}  Humans and objects are represented by nodes in the graph, while human-object interactions are represented by edges. In this experiment, we use a pre-trained  deformable convolutional network~\cite{dai2017deformable} for object detection and features extraction. Based on the detected bounding boxes, we extract node features ($7 \times 7 \times 80$) from the position-sensitive region of interest (PS RoI) pooling layer from the deformable ConvNet.
%We extract a $7 \times 7$ image feature for the detected class from the $7 \times 7 \times 80$ score maps (corresponding to 80 object categories).
We extract the edge feature from a combined bounding box, \ie, the smallest bounding box that contains both two nodes' bounding boxes.
% camera-ready: For edges between two nodes, we use ResNet~\cite{he2016deep} to extract the edge features by extracting the image feature for a combined bounding box, \ie, the smallest bounding box that contains both two nodes' bounding boxes. The ResNet is finetuned by action labels.
The functions of GPNN are implemented as follows. We use a convolutional network (128-128-1)-Sigmoid($\cdot$) with $1 \times 1$ kernels for the link function.
%We use the same architecture as the previous experiment for message functions and update functions: [FC($d_{V}$-$d_{V}$), FC($d_{E}$-$d_{E}$)] for message function and GRU($d_{V}$) for update function. The propagation step number $S$ is set to be 3. We use a FC($d_{V}$-117)-Sigmoid($\cdot$) for the readout function.
The message functions are composed of a fully connected layer, concatenation, and summation. For a node $v$, the neighboring node feature $\Gamma_{w}$ and edge feature $\Gamma_{vw}$ are passed through a fully connected layer and concatenated. The final incoming message is a weighted sum of messages from all neighboring nodes. Specifically, the message for node $v$ coming from node $w$ through edge $e=(v, w)$ is the concatenation of output from FC($d_{V}$-$d_{V}$) and FC($d_{E}$-$d_{E}$). A GRU($d_{V}$) is used for the update function. The propagation step number $S$ is set to be 3. For the readout function, we use a FC($d_{V}$-117)-Sigmoid($\cdot$) and FC($d_{V}$-26)-Sigmoid($\cdot$) for HICO-DET and V-COCO, respectively.

The probability of an HOI label of a human-object pair is given by the product of the final output probabilities from the human node and the object node.
% camera-ready: product of ... and connectivity
% The model detects the HOI with the highest probability across the possible human-object pairs in an image.
We employ an L1 loss for the adjacency matrix. For the node outputs, we use a weighted multi-class multi-label hinge loss. The reasons are two-folds: the training examples are not balanced, and it is essentially a multi-label problem for each node (there might not even exist a meaningful human-object interaction for detected humans and objects).

Our model is implemented using PyTorch and trained with a machine with a single Nvidia Titan Xp GPU. We start with a learning rate of 1e-3, and the rate decays every 5 epochs by 0.8. The training process takes about 20 epochs ($\sim$15 hours) to roughly converge with a batch size of 32.
% camera-ready: careful with large batch size

%\begin{table*}[t!]
%\begin{center}
%\caption{Comparison of constant graph flow and variable graph flow}
%
%\resizebox{0.9\textwidth}{!}{
%\begin{tabular}{c|ccc|cc|cc}
% \Xhline{2\arrayrulewidth}
% & \multicolumn{3}{c|}{HICO Detection mAP} & \multicolumn{2}{c|}{CAD-120 Detection F1-score} & \multicolumn{2}{c}{CAD-120 Prediction F1-score} \\
%\hline
%\textbf{Method} & Full & Rare & Non-rare& \makecell{Sub-activity(\%)} & Affordance(\%) & \makecell{Sub-activity(\%)} & Affordance(\%) \\
%\hline
%GPNN (constant graph) & 10.31 & 6.70 & 11.35 & 82.3 & 86.5 & 76.3 & 80.5 \\
%GPNN (parse graph) & 16.09 & 10.55 & 16.25 & 84.2 & 88.6 & 80.4 & 84.9 \\
% \Xhline{2\arrayrulewidth}
%\end{tabular}
%}
%\label{tab:graph_comparison}
%\end{center}
%\end{table*}

%%%%%%%%%%%%%%%%%%%%%%%%%%%%%Table 2%%%%%%%%%%%%%%%%%%%%%%%%%%%%%%%%%%%%%%%%%%%%
\begin{table}[t]%[!htbp]
\caption{\textbf{HOI detection results (mAP) on V-COCO~\cite{gupta2015visual}  dataset}. \textbf{Legend:} \textit{Set 1} indicates 18 HOI actions with one object, and \textit{Set 2} corresponds to 3 HOI actions (\ie, \emph{cut}, \emph{eat}, \emph{hit}) with two objects (\textit{instrument} and \textit{object}). %Higher values are better. The best scores are marked in \textbf{bold}
%See \autoref{sec:humanobjectinter} for more details
}
\label{tab:vcoco_result}
\resizebox{\textwidth}{!}{
\setlength\tabcolsep{5pt}
\renewcommand\arraystretch{1.1}
\begin{tabular}{l||c||c||c}  % {lccc}
\hline\thickhline
Method & Set 1 (mAP \%) $\uparrow$ & Set 2 (mAP \%) $\uparrow$ & Ave. (mAP \%) $\uparrow$ \\
\hline
\hline
 Gupta \etal \cite{gupta2015visual} &33.5 &26.7 &31.8\\
 InteractNet \cite{gkioxari2017detecting} &42.2 &33.2 &40.0\\
 \hline
\textbf{GPNN} &\textbf{44.5} &\textbf{42.8} &\textbf{44.0} \\
\textit{Performance Gain}(\%) 	& 5.5 & 28.9 & 10.0 \\
\hline
\end{tabular}
}
\end{table}

\begin{figure*}[t]
\centering
\includegraphics[width=\linewidth]{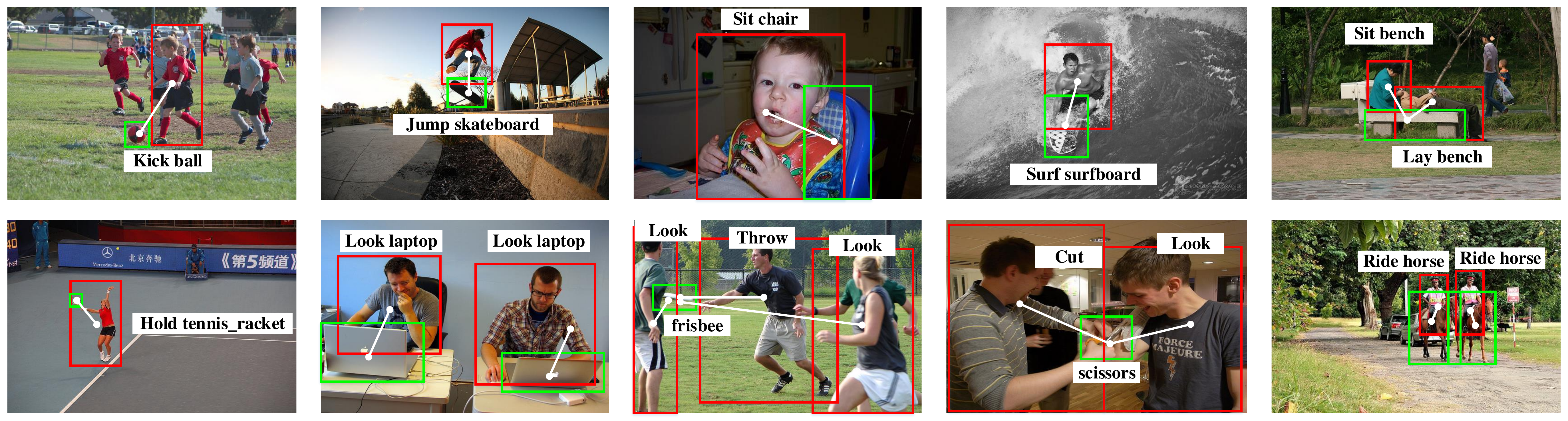}
\caption{\textbf{HOI detection results on V-COCO~\cite{gupta2015visual} test images.} Human and objects are shown in red and green rectangles, respectively. Best viewed in color.}
\label{fig:vcoco_results}
%\vspace{-10pt}
\end{figure*}

\noindent\textbf{Comparative Methods.} We compare our method with eight baselines: % which are mainly based on Fast-RCNN~\cite{girshick2015fast} and HO-RCNN~\cite{chao2017learning}. Fast-RCNN~\cite{girshick2015fast} has been shown as an effective network architecture for object detection. HO-RCNN is a strong baseline since it is a modified RCNN architecture which is specifically designed for the HOI detection task. Specially, the comparative methods are:
(1) Fast-RCNN (union)~\cite{girshick2015fast}: for each human-object
proposal from detection results, their attention windows are used as the region proposal for Fast-RCNN. (2) Fast-RCNN (score)~\cite{girshick2015fast}: given human-object proposals, HOI is predicted by linearly combining the human and object detection scores. (3) HO-RCNN~\cite{chao2017learning}: a multi-stream architecture with a ConvNet to classify human, object and human-object proposals, respectively. The final output is computed by combining the scores from all the three streams. (4) HO-RCNN+IP~\cite{chao2017learning} and (5) HO-RCNN+IP+S~\cite{chao2017learning}:  HO-RCNN with additional components. Interaction Patterns (IP) acts as a attention filter to images. S is an extra path with a single neuron that uses the raw object detection score to produce an offset for the final detection. More detailed descriptions of above five baselines can be found in~\cite{chao2017learning}. (6) Gupta \etal \cite{gupta2015visual}: trained based on Fast-RCNN \cite{girshick2015fast}. We use the scores reported in \cite{gkioxari2017detecting}.
(7) Shen \etal \cite{shenscaling}: final predictions are from two Faster RCNN \cite{ren2015faster} based networks which are trained for predicting verb and object classes, respectively. (8) InteractNet \cite{gkioxari2017detecting}: a modified Faster RCNN \cite{ren2015faster} with an additional human-centric branch that estimates an action-specific density map for locating objects.

\noindent\textbf{Experiment Results.} Following the standard settings in HICO-DET and V-COCO  benchmarks, we evaluate HOI detection using mean average precision (mAP). An HOI detection is considered as a true positive when the human detection, the object detection, and the interaction class are all correct. The human and object bounding boxes are considered as true positives if they overlap with a ground truth bounding boxes of the same class with an intersection over union (IoU) greater than 0.5. For HICO-DET dataset, we report the mAP over three different HOI category sets: i) all 600 HOI categories in HICO (Full); ii) 138 HOI categories with less than 10 training instances (Rare); and iii) 462 HOI categories with 10 or more training instances (Non-Rare). For V-COCO dataset, since we concentrate on HOI detection, we report the mAP on three groups: i) 18 HOI action classes with one target object; ii) 3 HOI categories with two types of objects; iii) all 24 (=$18+3\times2$) HOI classes.
Results are evaluated on the test sets and reported in \autoref{tab:hico_result} and \autoref{tab:vcoco_result}.

As shown in \autoref{tab:hico_result}, the proposed GPNN substantially outperforms the comparative methods, achieving \textbf{31.89\%}, \textbf{30.45\%}, and \textbf{32.13\%} improvement over the second best methods on the three HOI category sets on the HICO-DET dataset. The results on V-COCO dataset (in \autoref{tab:vcoco_result}) also consistently demonstrate the superior performance of the proposed GPNN.
 %We almost double the accuracy on all three HOI category sets.
Two important \textbf{conclusions} can be drawn from the results:
\textbf{i)} our method  is scalable to large datasets; \textbf{ii)} and our method performs better than pure neural network.
%\vspace{-2mm}
%\begin{itemize}
%\item[\textbf{i)}] Our method  is scalable to large datasets.
%\item[\textbf{ii)}] Our method performs better than pure neural network approaches, which well demonstrates the benefit of leveraging structure information in HOI domain.
%\end{itemize}
%\vspace{-2mm}
Some visual results can be found in \autoref{fig:hico_results} and \autoref{fig:vcoco_results}.

%\begin{figure}[t]
%\centering
%\subfloat[Action]{ \includegraphics[trim={0.6in 0.2in 1.2in 0.15in}, clip, width=0.48\linewidth]{../fig/raw/cad_results/prediction/confusion_subactivity.pdf} }
%\subfloat[Affordance]{ \includegraphics[trim={0.57in 0.2in 1.2in 0.15in}, clip, width=0.48\linewidth]{../fig/raw/cad_results/prediction/confusion_affordance.pdf} }
%\caption{Confusion matrices of prediction results on CAD-120.}
%\label{fig:cad_prediction_confusion}
%\end{figure}

%\begin{figure*}[h!]
%\centering
%\includegraphics[width=\linewidth]{../fig/cad_qualitative.pdf}
%\caption{Qualitative results of a ``microwaving food'' activity on CAD-120. Detection and prediction results are shown by different bars. The label of the sub-activity at time $t$ is predicted at time $t\text{-}1$.}
%\label{fig:cad_qualitative}
%\end{figure*}

\subsection{Human-Object Interaction Recognition in Videos}
\label{sec:human_activity_recognition}
The goal of this experiment is to detect and predict the human sub-activity labels and object affordance labels as the human-object interaction progresses in videos. The problem is challenging since it involves complex interactions that humans make with multiple objects, and objects also interact with each other.

\noindent\textbf{CAD-120 dataset~\cite{koppula2016anticipating}.} %We perform experiments on the CAD-120 dataset~\cite{koppula2016anticipating}.
It has 120 RGB-D videos of 4 subjects performing 10 activities, each of which is a sequence of sub-activities involving 10 actions (\eg, reaching, opening), and 12 object affordances (\eg, reachable, openable) in total.

%%%%%%%%%%%%%%%%%%%%%%%%%%%%%Table 2%%%%%%%%%%%%%%%%%%%%%%%%%%%%%%%%%%%%%%%%%%%%
\begin{table}[t]%[!htbp]
\caption{\textbf{Human activity detection and future anticipation results on CAD-120~\cite{koppula2016anticipating} dataset}, measured via F1-score. %Higher values are better. The best scores are marked in \textbf{bold}
%See \autoref{sec:human_activity_recognition} for more details
}
\label{tab:cad120_result}
\resizebox{\textwidth}{!}{
\setlength\tabcolsep{5pt}
\renewcommand\arraystretch{1.1}
\begin{tabular}{l||c|c||c|c}  % {lccc}
\hline\thickhline
 & \multicolumn{2}{c||}{Detection (F1-score) $\uparrow$} & \multicolumn{2}{c}{Anticipation (F1-score) $\uparrow$} \\
\cline{2-5}
 Method&\makecell{Sub\\-activity(\%)} &\makecell{Object\\Affordance(\%)} &\makecell{Sub\\-activity(\%)} &\makecell{Object\\Affordance(\%)}\\
\hline
\hline
 ATCRF~\cite{koppula2016anticipating} & 80.4 & 81.5 & 37.9 & 36.7 \\
 S-RNN~\cite{jain2016structural} & 83.2 & 88.7 & 62.3 & 80.7 \\
 S-RNN (multi-task)~\cite{jain2016structural} & 82.4 & \textbf{91.1} & 65.6 & 80.9 \\
\hline
\textbf{GPNN} &\textbf{88.9} &88.8 &\textbf{75.6} &\textbf{81.9}\\
%\textit{Performance Gain} 	& +1.0 & - & +14.8 & +4.0 \\
\textit{Performance Gain}(\%) 	& 8.1 & - & 15.2 & 1.2 \\
\hline
\end{tabular}
}
\end{table}

\noindent\textbf{Implementation Details.} The link function is implemented as: convLSTM(1024-1024-1024-1)-Sigmoid($\cdot$) (\ie, a four-layer convLSTM). We use the same architecture as the previous experiment for message functions and update functions: [FC($d_{V}$-$d_{V}$), FC($d_{E}$-$d_{E}$)] for message function and GRU($d_{V}$) for update function. The propagation step number $S$ is set to be 3.
We use a FC($d_{V}$-10)-Softmax($\cdot$) and a FC($d_{V}$-12)-Softmax($\cdot$) for readout functions of sub-activity and object affordance detection/anticipation, respectively. We employ an L1 loss for the adjacency matrix and a cross entropy loss for the node outputs. We use the publicly available node and edge features from \cite{koppula2013learning}.

\begin{figure*}[t]
\centering
\includegraphics[width=\linewidth]{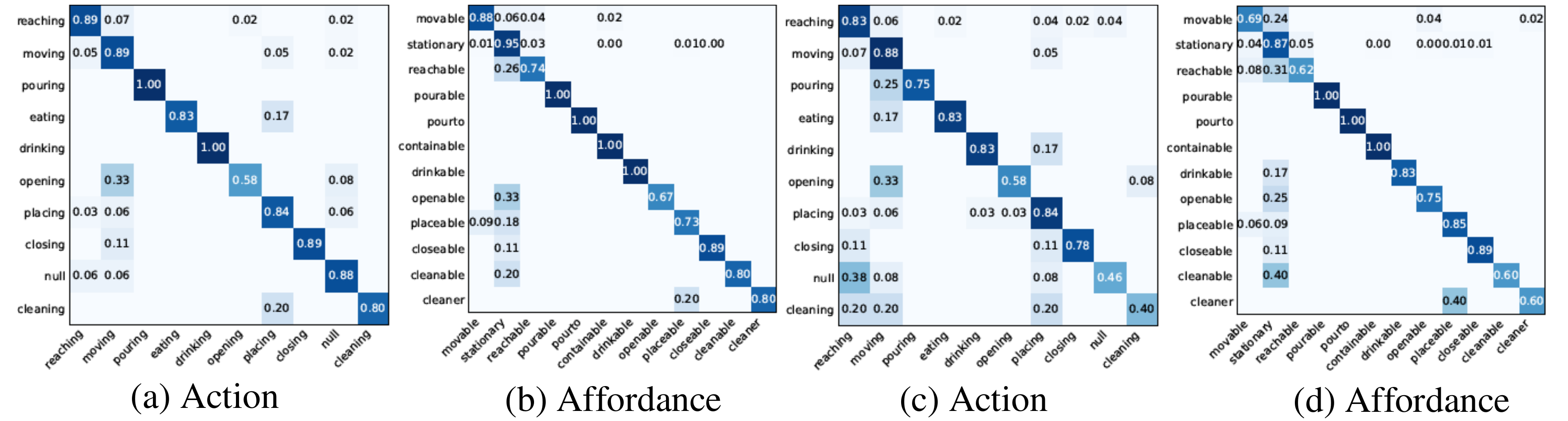}
\caption{\textbf{Confusion matrices of HOI detection (a)(b) and anticipation (c)(d) results on CAD-120~\cite{koppula2016anticipating} dataset.} Zoom in for more details.}
\label{fig:cad_detection_confusion}
%\vspace{-10pt}
\end{figure*}

%\begin{figure}[t]
%\centering
%\subfloat[Action]{\includegraphics[trim={41pt 0pt 75pt 0pt},clip,width=0.32\linewidth]{figs/iteration1.pdf}}
%\subfloat[Action]{\includegraphics[trim={41pt 0pt 75pt 0pt},clip,width=0.32\linewidth]{figs/iteration2.pdf}}
%\subfloat[Action]{\includegraphics[trim={41pt 0pt 75pt 0pt},clip,width=0.32\linewidth]{figs/iteration3.pdf}}
%\vspace{-10pt}
%\caption{\textbf{Confusion matrices of detection (a)(b) and prediction (c)(d) results on CAD-120~\cite{koppula2016anticipating} dataset.} Zoom in for more details.}
%\vspace{-10pt}
%\label{fig:cad_detection_confusion}
%\end{figure}

%\begin{table}
%\begin{center}
%\resizebox{0.48\textwidth}{!}{
%\begin{tabular}{c|cc|cc}
% \Xhline{2\arrayrulewidth}
% & \multicolumn{2}{c|}{Detection F1-score} & \multicolumn{2}{c}{Prediction F1-score} \\
% \Xhline{2\arrayrulewidth}
% \textbf{Method} & \makecell{Sub-\\activity(\%)} & Affordance(\%) & \makecell{Sub-\\activity(\%)} & Affordance(\%) \\
% \hline
% ATCRF~\cite{koppula2016anticipating} & 80.4 & 81.5 & 37.9 & 36.7 \\
% S-RNN~\cite{jain2016structural} & 83.2 & 88.7 & 62.3 & 80.7 \\
% S-RNN (multi-task)~\cite{jain2016structural} & 82.4 & \textbf{91.1} & 65.6 & 80.9 \\
% GPNN & \textbf{84.2} & 88.6 & \textbf{80.4} & \textbf{84.9} \\
% \Xhline{2\arrayrulewidth}
% 	& +1.0 & - & +14.8 & +4.0 \\
%\end{tabular}
%}
%\caption{Detection and future prediction results on the CAD-120 dataset}
%\label{tab:cad120_result}
%\end{center}
%\end{table}

\noindent\textbf{Comparative Methods.}  We compare our method with two baselines: anticipatory temporal CRF (ATCRF)~\cite{koppula2016anticipating} and structural RNN (S-RNN)~\cite{jain2016structural}. ATCRF is a top-performing graphical model approach for this problem, while S-RNN is the state-of-art method using structured neural networks. ATCRF models the human activities through a spatial-temporal conditional random field. S-RNN casts a pre-defined spatial-temporal graph as an RNN mixture by representing nodes and edges as LSTMs.

\begin{figure*}[t]
\centering
\includegraphics[width=\linewidth]{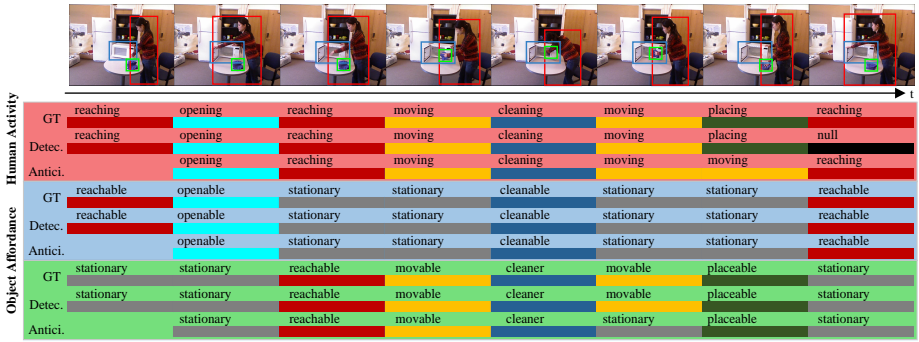}
%\vspace{-15pt}
\caption{\textbf{HOI detection results on a ``cleaning objects'' activity} on CAD-120~\cite{koppula2016anticipating} dataset. Human are shown in red rectangle. Two objects are shown in green and blue rectangles, respectively. Detection and anticipation results are shown by different bars. For anticipation task, the label of the sub-activity at time $t$ is anticipated at time $t\text{-}1$. %Zoom-in for details.
}
\label{fig:cad_qualitative}
%\vspace{-10pt}
\end{figure*}

\noindent\textbf{Experiment Results.} In \autoref{tab:cad120_result} we show the quantitative comparison of our method with other competitors. It shows the F1-scores averaged over all classes on detection and activity anticipation tasks. GPNN greatly improves over ATCRF and S-RNN, especially on anticipation task. %It is important to note that all three methods use the same graph structure as shown in Fig.~\ref{fig:overview} (b).
Our method outperforms the other two for the following reasons. i) Comparing to ATCRF limited to the Markov assumption, our method allows arbitrary graph structures with improved representation ability. ii) Our method enjoys the benefit of deep integration of graphical models and neural networks and can be learned in an end-to-end manner. iii) Rather than relying on a pre-fixed graph structure as in S-RNN, we infer the graph structure via learning an adjacency matrix and thus be able to control the information flow between nodes during massage passing. \autoref{fig:cad_detection_confusion} show the confusion matrices for detecting and predicting the sub-activities and object affordances, respectively. From above results we can draw two important \textbf{conclusions}:
\textbf{i)} our method is well applicable to the spatio-temporal domain; and \textbf{ii)} our method outperforms pure graphical models (\eg, ATCRF) and deep networks with pre-fixed graph structures (\eg, S-RNN).
%\vspace{-2mm}
%\begin{itemize}
%\item[\textbf{i)}] our method is well applicable to complex HOI settings in spatiotemporal domain; and
%\item[\textbf{ii)}] our method is more favored than graph models (\eg, ATCRF) and deep networks with pre-fixed graph structures (\eg, S-RNN).
%\end{itemize}
%\vspace{-2mm}
\autoref{fig:cad_qualitative} shows a qualitative visualization of ``cleaning objects''. We show one representative frame for each sub-activity as well as the corresponding detections and anticipations.

%\begin{figure}[t]
%\centering
%\includegraphics[width=\linewidth]{../fig/hico_example.pdf}
%\caption{An example of the human-object interaction detection task. From a static image, humans and objects are detected to construct the complete graph. The final parse graph with an HOI label is given by the GPNN.}
%\label{fig:hico_example}
%\end{figure}

%\begin{table}
%\begin{center}
%\caption{Detection results (mAP) on the HICO-DET dataset}
%\resizebox{0.48\textwidth}{!}{
%\begin{tabular}{l|c|c|c}
% \Xhline{2\arrayrulewidth}
% \textbf{Method} & Full & Rare & Non-rare \\
% \hline
% Random & $1.35\times10^{-3}$ & $5.72\times10^{-4}$ & $1.62\times10^{-3}$ \\
% Fast-RCNN(union)~\cite{girshick2015fast} & 1.75 & 0.58 & 2.10 \\
% Fast-RCNN(score)~\cite{girshick2015fast} & 2.85 & 1.55 & 3.23 \\
% HO-RCNN~\cite{chao2017learning} & 5.73 & 3.21 & 6.48 \\
% HO-RCNN+IP+S~\cite{chao2017learning} & 7.81 & 5.37 & 8.54 \\
% GPNN & \textbf{16.09} & \textbf{10.55} & \textbf{16.25} \\
% \Xhline{2\arrayrulewidth}
% 	& +8.28 & +5.18 & +7.71 \\
%\end{tabular}
%}
%\label{tab:hico_result}
%\end{center}
%\end{table}

\subsection{Ablation Study}
\label{sec:abastudy}
In this section, we analyze the contributions of different model components to the final performance and examine the effectiveness of our main assumptions. \autoref{tab:graph_comparison} shows the detailed results on all three datasets.

\noindent\textbf{Integration of DNN with Graphical Model.} We first examine the influence of integrating DNN with a graphical model. We directly feed the features, which are originally used for GPNN, into different fully connected networks for predicting HOI action or object classes. From \autoref{tab:graph_comparison}, we can observe the performance of \textit{w/o graph} is significantly worse than GPNN model over various HOI datasets. This supports our view that modeling high-level structures and leveraging learning capabilities of DNNs together is essential for HOI tasks.

\noindent\textbf{GPNN with Fixed Graph Structures.} In \autoref{sec:gpnn}, GPNN automatically infers graph structures (\ie, parse graph) via learning a soft adjacency matrix. To assess this strategy, we fix all the entries in the soft adjacency matrices to be constant 1. This way the graph structures are fixed and the information flow between nodes are not weighted. For \textit{constant graph} baseline, we see obvious performance decrease, compared with the full GPNN model. This indicates that inferring graph structures is critical to get reasonable performance.

\noindent\textbf{GPNN without Supervision on Link Functions.} We perform experiments by turning off the L1 loss on adjacency matrices (\textit{w/o graph loss} in \autoref{tab:graph_comparison}). We can observe that the intermediate L1 loss is effective, further verifying our design to learn the graph structure. Another interesting observation is that training the model without this loss has a similar effect to training with constant graph. Hence supervision on the graph is fairly important.

\noindent\textbf{Jointly Learning Parse Graph and Message Passing.} We next study the effect of jointly learning graph structures and message passing. By isolating graph parsing from message passing, we obtain \textit{w/o joint parsing}, where the adjacency matrices are directly computed by link functions from edge features at the beginning. We observe a performance decrease in \autoref{tab:graph_comparison}, showing that learning graph structures and message passing together indeed boost the performance.

\noindent\textbf{Iterative Learning Process.} Next we examine the effect of iterative message passing, we report three baselines: \textit{1 iteration}, \textit{2 iterations}, and \textit{4 iterations}, which correspond to the results from different message passing iterations. The baseline \textit{GPNN} (first row in \autoref{tab:graph_comparison}) are the results after three iterations. From the results we observe that the iterative learning process is able to gradually improve the performance in general. We also observe that when the iteration round is increased to a certain extent, the performance drops slightly.
\begin{table}[t]%[!htbp]
\caption{\textbf{Ablation study of GPNN model.} Higher values are better.
%See \autoref{sec:abastudy} for more details
}
\resizebox{\textwidth}{!}{
\setlength\tabcolsep{2pt}
\renewcommand\arraystretch{1.1}
\begin{tabular}{l|l||c|c|c|c|c|c|c|c|c|c}  % {lccc}
\hline\thickhline
 &&  \multicolumn{3}{c|}{V-COCO~\cite{gupta2015visual}} & \multicolumn{3}{c|}{HICO-DET~\cite{chao2017learning}} & \multicolumn{4}{c}{CAD-120~\cite{koppula2016anticipating}}\\
\cline{3-12}
 &&  \multicolumn{3}{c|}{\makecell{HOI Detection \\mAP(\%) $\uparrow$}} & \multicolumn{3}{c|}{\makecell{HOI Detection \\mAP(\%) $\uparrow$}} & \multicolumn{2}{c|}{\makecell{HOI Detec. \\ F1-score(\%) $\uparrow$}} & \multicolumn{2}{c}{\makecell{HOI Antici. \\F1-score(\%) $\uparrow$}} \\
\cline{3-12}
Aspect &Method & Set 1 &Set 2 &Ave. & Full & Rare & \makecell{Non-\\rare}& \makecell{Sub-\\activity} & \makecell{Object\\Aff.(\%)} & \makecell{Sub-\\activity} &\makecell{Object\\Aff.(\%)}\\
\hline
\hline
&\makecell{GPNN\\(3 iterations)} & \textbf{44.5}  & \textbf{42.8}  & \textbf{44.0}  & \textbf{13.11} & \textbf{9.34} & \textbf{14.23} & \textbf{88.9} & \textbf{88.8} & 75.6 & \textbf{81.9} \\
\hline
\hline
& w/o graph   & 27.4  & 30.0  & 28.1  & 7.88 & 2.04 & 9.62 & 50.2 & 20.8 & 32.3 & 19.6 \\
\textit{graph }& constant graph & 34.6  & 33.3  & 34.3  & 8.75 & 1.94 & 10.79 & 85.3 & 85.6 & 73.8 & 79.1 \\
\textit{structure} & w/o graph loss & 37.7  & 40.5  & 38.4  & 8.15  & 6.24  & 8.72  & 85.2 & 85.8 & 74.7 & 79.2 \\
& w/o joint parsing & 43.6  & 39.4  & 42.5  & 10.17  & 5.81  & 11.47  & 79.3 & 79.2 & 74.7 & 80.3 \\
\hline
\textit{iterative} & 1 iteration & 42.0  & 40.7  & 41.7  & 11.38 & 7.27 & 12.61 & 80.5 & 80.7 & 75.2 & 81.1 \\
\textit{learning} & 2 iterations & 44.1  & 42.2  & 43.6  & 12.37 & 9.01 & 13.38 & 87.9 & 86.1 & \textbf{76.1} & 81.5 \\
&4 iterations & 43.6  & 40.9  & 42.9  & 12.39 & 8.95 & 13.41 & 87.9 & 85.7 & 75.5 & 80.6 \\
\hline
\end{tabular}
}
\label{tab:graph_comparison}
\end{table}

\section{Conclusion}
In this paper, we propose Graph Parsing Neural Network (GPNN) for inferring a parse graph in an end-to-end manner. The network can be decomposed into four distinct functions, namely link functions, message functions, update functions and readout functions, for iterative graph inference and message passing. GPNN provides a generic HOI representation that is applicable in both spatial and spatial-temporal domains. We demonstrate a substantial performance gain on three HOI datasets, showing the effectiveness of the proposed framework.\\
%: HICO-DET~\cite{chao2017learning}, V-COCO~\cite{gupta2015visual} and CAD-120~\cite{koppula2016anticipating}, we qualitatively and quantitatively show the effectiveness of GPNN by demonstrating a significant performance gain on different HOI tasks. The experimental results also clearly demonstrate the effectiveness of the proposed joint parsing framework.

\noindent\textbf{Acknowledgments.}
The authors thank Prof. Ying Nian Wu from UCLA Statistics Department for helpful comments on this work.
This research is supported by DARPA XAI N66001-17-2-4029, ONR MURI N00014-16-1-2007, ARO W911NF1810296, and N66001-17-2-3602.

{\small
\bibliographystyle{splncs04}
\bibliography{paper}

\begin{thebibliography}{10}
\providecommand{\url}[1]{\texttt{#1}}
\providecommand{\urlprefix}{URL }
\providecommand{\doi}[1]{https://doi.org/#1}

\bibitem{chao2017learning}
Chao, Y.W., Liu, Y., Liu, X., Zeng, H., Deng, J.: Learning to detect
  human-object interactions (2018)

\bibitem{chao2015hico}
Chao, Y.W., Wang, Z., He, Y., Wang, J., Deng, J.: {HICO}: A benchmark for
  recognizing human-object interactions in images. In: ICCV (2015)

\bibitem{chen2016deeplab}
Chen, L.C., Papandreou, G., Kokkinos, I., Murphy, K., Yuille, A.L.: Deeplab:
  Semantic image segmentation with deep convolutional nets, atrous convolution,
  and fully connected {CRF}s. PAMI  (2016)

\bibitem{chen2015learning}
Chen, L.C., Schwing, A., Yuille, A., Urtasun, R.: Learning deep structured
  models. In: ICML (2015)

\bibitem{cho2014properties}
Cho, K., Van~Merri{\"e}nboer, B., Bahdanau, D., Bengio, Y.: On the properties
  of neural machine translation: Encoder--decoder approaches. Syntax, Semantics
  and Structure in Statistical Translation p.~103 (2014)

\bibitem{dai2017deformable}
Dai, J., Qi, H., Xiong, Y., Li, Y., Zhang, G., Hu, H., Wei, Y.: Deformable
  convolutional networks. In: ICCV (2017)

\bibitem{defferrard2016convolutional}
Defferrard, M., Bresson, X., Vandergheynst, P.: Convolutional neural networks
  on graphs with fast localized spectral filtering. In: NIPS (2016)

\bibitem{delaitre2011learning}
Delaitre, V., Sivic, J., Laptev, I.: Learning person-object interactions for
  action recognition in still images. In: NIPS (2011)

\bibitem{desai2012detecting}
Desai, C., Ramanan, D.: Detecting actions, poses, and objects with relational
  phraselets. In: ECCV (2012)

\bibitem{elman1990finding}
Elman, J.L.: Finding structure in time. Cognitive science  (1990)

\bibitem{Fang2018}
Fang, H.S., Xu, Y., Wang, W., Zhu, S.C.: Learning pose grammar to encode human
  body configuration for 3d pose estimation. In: AAAI (2018)

\bibitem{gilmer2017neural}
Gilmer, J., Schoenholz, S.S., Riley, P.F., Vinyals, O., Dahl, G.E.: Neural
  message passing for quantum chemistry. In: ICML (2017)

\bibitem{girshick2015fast}
Girshick, R.: Fast {R-CNN}. In: ICCV (2015)

\bibitem{gkioxari2017detecting}
Gkioxari, G., Girshick, R., Doll{\'a}r, P., He, K.: Detecting and recognizing
  human-object interactions. In: CVPR (2018)

\bibitem{gupta2007objects}
Gupta, A., Davis, L.S.: Objects in action: An approach for combining action
  understanding and object perception. In: CVPR (2007)

\bibitem{gupta2009observing}
Gupta, A., Kembhavi, A., Davis, L.S.: Observing human-object interactions:
  Using spatial and functional compatibility for recognition. PAMI  (2009)

\bibitem{gupta2015visual}
Gupta, S., Malik, J.: Visual semantic role labeling. arXiv preprint
  arXiv:1505.04474  (2015)

\bibitem{hochreiter1997long}
Hochreiter, S., Schmidhuber, J.: Long short-term memory. Neural computation
  (1997)

\bibitem{hu2013recognising}
Hu, J.F., Zheng, W.S., Lai, J., Gong, S., Xiang, T.: Recognising human-object
  interaction via exemplar based modelling. In: ICCV (2013)

\bibitem{jain2016structural}
Jain, A., Zamir, A.R., Savarese, S., Saxena, A.: {Structural-RNN}: Deep
  learning on spatio-temporal graphs. In: CVPR (2016)

\bibitem{kipf2017semi}
Kipf, T.N., Welling, M.: Semi-supervised classification with graph
  convolutional networks. In: ICLR (2017)

\bibitem{koppula2016anticipating}
Koppula, H.S., Saxena, A.: Anticipating human activities using object
  affordances for reactive robotic response. PAMI  (2016)

\bibitem{koppula2013learning}
Koppula, H.S., Gupta, R., Saxena, A.: Learning human activities and object
  affordances from {RGB-D} videos. The International Journal of Robotics
  Research  (2013)

\bibitem{li2017situation}
Li, R., Tapaswi, M., Liao, R., Jia, J., Urtasun, R., Fidler, S.: Situation
  recognition with graph neural networks. In: ICCV (2017)

\bibitem{li2015gated}
Li, Y., Tarlow, D., Brockschmidt, M., Zemel, R.: Gated graph sequence neural
  networks. In: ICLR (2016)

\bibitem{liang2017interpretable}
Liang, X., Lin, L., Shen, X., Feng, J., Yan, S., Xing, E.P.: Interpretable
  structure-evolving lstm. In: ICCV (2017)

\bibitem{liang2016semantic}
Liang, X., Shen, X., Feng, J., Lin, L., Yan, S.: Semantic object parsing with
  graph lstm. In: ECCV (2016)

\bibitem{lin2014microsoft}
Lin, T.Y., Maire, M., Belongie, S., Hays, J., Perona, P., Ramanan, D.,
  Doll{\'a}r, P., Zitnick, C.L.: Microsoft coco: Common objects in context. In:
  ECCV (2014)

\bibitem{mallya2016learning}
Mallya, A., Lazebnik, S.: Learning models for actions and person-object
  interactions with transfer to question answering. In: ECCV (2016)

\bibitem{marino2016more}
Marino, K., Salakhutdinov, R., Gupta, A.: The more you know: Using knowledge
  graphs for image classification. In: CVPR (2016)

\bibitem{monti2016geometric}
Monti, F., Boscaini, D., Masci, J., Rodol{\`a}, E., Svoboda, J., Bronstein,
  M.M.: Geometric deep learning on graphs and manifolds using mixture model
  cnns. CVPR  (2016)

\bibitem{niepert2016learning}
Niepert, M., Ahmed, M., Kutzkov, K.: Learning convolutional neural networks for
  graphs. In: ICML (2016)

\bibitem{park2017attribute}
Park, S., Nie, X., Zhu, S.C.: Attribute and-or grammar for joint parsing of
  human pose, parts and attributes. PAMI  (2017)

\bibitem{qi2017predicting}
Qi, S., Huang, S., Wei, P., Zhu, S.C.: Predicting human activities using
  stochastic grammar. In: ICCV (2017)

\bibitem{qi2018generalized}
Qi, S., Jia, B., Zhu, S.C.: Generalized earley parser: Bridging symbolic
  grammars and sequence data for future prediction. In: ICML (2018)

\bibitem{ren2015faster}
Ren, S., He, K., Girshick, R., Sun, J.: Faster {R-CNN}: Towards real-time
  object detection with region proposal networks. In: NIPS (2015)

\bibitem{seo2016structured}
Seo, Y., Defferrard, M., Vandergheynst, P., Bresson, X.: Structured sequence
  modeling with graph convolutional recurrent networks. arXiv preprint
  arXiv:1612.07659  (2016)

\bibitem{shenscaling}
Shen, L., Yeung, S., Hoffman, J., Mori, G., Fei-Fei, L.: Scaling human-object
  interaction recognition through zero-shot learning (2018)

\bibitem{xingjian2015convolutional}
Shi, X., Chen, Z., Wang, H., Yeung, D.Y., Wong, W.K., Woo, W.c.: Convolutional
  {LSTM} network: A machine learning approach for precipitation nowcasting. In:
  NIPS (2015)

\bibitem{simonovsky2017dynamic}
Simonovsky, M., Komodakis, N.: Dynamic edge-conditioned filters in
  convolutional neural networks on graphs. CVPR  (2017)

\bibitem{teney2016graph}
Teney, D., Liu, L., Hengel, A.v.d.: Graph-structured representations for visual
  question answering. In: CVPR (2017)

\bibitem{tompson2014joint}
Tompson, J.J., Jain, A., LeCun, Y., Bregler, C.: Joint training of a
  convolutional network and a graphical model for human pose estimation. In:
  NIPS (2014)

\bibitem{Wang_2018_CVPR}
Wang, W., Xu, Y., Shen, J., Zhu, S.C.: Attentive fashion grammar network for
  fashion landmark detection and clothing category classification. In: CVPR
  (2018)

\bibitem{wu2016deep}
Wu, Z., Lin, D., Tang, X.: Deep markov random field for image modeling. In:
  ECCV (2016)

\bibitem{xia2016pose}
Xia, F., Zhu, J., Wang, P., Yuille, A.L.: Pose-guided human parsing by an
  {And/Or} graph using pose-context features. In: AAAI (2016)

\bibitem{xu2017scene}
Xu, D., Zhu, Y., Choy, C.B., Fei-Fei, L.: Scene graph generation by iterative
  message passing. In: ICCV (2017)

\bibitem{yao2010grouplet}
Yao, B., Fei-Fei, L.: Grouplet: A structured image representation for
  recognizing human and object interactions. In: CVPR (2010)

\bibitem{yao2010modeling}
Yao, B., Fei-Fei, L.: Modeling mutual context of object and human pose in
  human-object interaction activities. In: CVPR (2010)

\bibitem{yao2011human}
Yao, B., Jiang, X., Khosla, A., Lin, A.L., Guibas, L., Fei-Fei, L.: Human
  action recognition by learning bases of action attributes and parts. In: ICCV
  (2011)

\bibitem{yuan2017temporal}
Yuan, Y., Liang, X., Wang, X., Yeung, D.Y., Gupta, A.: Temporal dynamic graph
  {LSTM} for action-driven video object detection. In: ICCV (2017)

\bibitem{zheng2015conditional}
Zheng, S., Jayasumana, S., Romera-Paredes, B., Vineet, V., Su, Z., Du, D.,
  Huang, C., Torr, P.H.: Conditional random fields as recurrent neural
  networks. In: ICCV (2015)

\end{thebibliography}
}

\end{document}